\pdfoutput=1
\documentclass{article}

    \PassOptionsToPackage{numbers, compress}{natbib}

\usepackage[preprint]{neurips_2024}

\usepackage[utf8]{inputenc} 
\usepackage[T1]{fontenc}    
\usepackage{url}            
\usepackage{booktabs}       
\usepackage{amsfonts}       
\usepackage{nicefrac}       
\usepackage{microtype}      
\usepackage{xcolor}         
\usepackage{multirow}
\usepackage{multicol}
\usepackage{amssymb}
\usepackage[ruled,vlined]{algorithm2e}
\usepackage{colortbl}
\usepackage{longtable}
\usepackage{mdframed}
\usepackage{inconsolata}
\usepackage{color}
\usepackage{amsmath}

\definecolor{darkred}{rgb}{0.8, 0.0, 0.0}

\usepackage{array}
\usepackage{graphicx}
\usepackage{caption}

\usepackage{float}
\usepackage{adjustbox}

\usepackage{cancel}

\usepackage[most]{tcolorbox}

\definecolor{citecolor}{HTML}{0071BC}
\definecolor{linkcolor}{HTML}{ED1C24}
\usepackage[pagebackref=true,breaklinks=true,colorlinks,citecolor=citecolor, linkcolor=linkcolor]{hyperref}

\definecolor{fontgray}{RGB}{44, 62, 80}
\definecolor{myred}{RGB}{235, 47, 6} 
\definecolor{summertime}{RGB}{245, 205, 121}
\definecolor{darkgrass}{RGB}{0, 148, 50}
\definecolor{myblue}{RGB}{0, 168, 255}
\definecolor{mygray}{RGB}{158, 158, 158}
\definecolor{puffin}{RGB}{250, 152, 58}
\definecolor{lowpurple}{RGB}{210, 180, 222}
\definecolor{lowblue}{RGB}{102,178,255}
\definecolor{lowred}{RGB}{245, 183, 177}
\definecolor{deeppurple}{RGB}{142, 68, 173}
\definecolor{nephritis}{RGB}{39, 174, 96}

\definecolor{deepblue}{RGB}{41, 128, 185}
\definecolor{shymoment}{RGB}{162, 155, 254}
\definecolor{firstdate}{RGB}{250, 177, 160}
\definecolor{mintleaf}{RGB}{0, 184, 148}
\definecolor{alizarin}{RGB}{231, 76, 60}
\definecolor{soaring}{RGB}{149, 175, 192}
\definecolor{electronblue}{RGB}{9, 132, 227}
\definecolor{pinkgla}{RGB}{0, 184, 148}
\definecolor{coral}{RGB}{255, 127, 80}

\newcommand{\squishlist}{
\begin{list}{$\bullet$}
{   \setlength{\itemsep}{0pt}
   \setlength{\parsep}{3pt}
   \setlength{\topsep}{3pt}
   \setlength{\partopsep}{0pt}
   \setlength{\leftmargin}{1.5em}
   \setlength{\labelwidth}{1em}
   \setlength{\labelsep}{0.5em} } }
\newcounter{Lcount}
\newcommand{\squishlisttwo}{
\begin{list}{\arabic{Lcount}. }
  { \usecounter{Lcount}
 \setlength{\itemsep}{0pt}
 \setlength{\parsep}{0pt}
 \setlength{\topsep}{0pt}
 \setlength{\partopsep}{0pt}
 \setlength{\leftmargin}{2em}
 \setlength{\labelwidth}{1.5em}
 \setlength{\labelsep}{0.5em} } }
\newcommand{\squishend}{\end{list} }

\definecolor{lightpink}{rgb}{0.945, 0.816, 0.804}
\definecolor{lightgreen}{rgb}{0.851, 0.906, 0.839}
\definecolor{lightblue}{rgb}{0.8, 0.9, 1}
\definecolor{lightyellow}{rgb}{0.992, 0.949, 0.816}

\newtcolorbox[list inside=prompt,auto counter,number within=section]{prompt}[1][]{
    colbacktitle=black!60,
    coltitle=white,
    fontupper=\footnotesize,
    boxsep=5pt,
    enhanced,
    left=0pt,
    right=0pt,
    top=0pt,
    bottom=0pt,
    boxrule=1pt,
    breakable,
    #1
}

\title{HuatuoGPT-o1, Towards Medical Complex Reasoning with LLMs}

\author{Junying Chen$^{1}$, Zhenyang Cai$^{1}$, Ke Ji$^{1}$, Xidong Wang$^{1}$, Wanlong Liu$^{1}$ \\
\textbf{Rongsheng Wang}$^{1}$, \textbf{Jianye Hou}$^{1}$, \textbf{Benyou Wang}$^{1,2}$\thanks{Benyou is the corresponding author with email: \textit{wangbenyou@cuhk.edu.cn}} \\
$^1$ The Chinese University of Hong Kong, Shenzhen \\
$^2$ Shenzhen Research Institute of Big Data \\
\url{https://github.com/FreedomIntelligence/HuatuoGPT-o1}}

\begin{document}

\maketitle

\begin{abstract}
The breakthrough of OpenAI o1 highlights the potential of enhancing reasoning to improve LLM. Yet, most research in reasoning has focused on mathematical tasks, leaving domains like medicine underexplored.  The medical domain, though distinct from mathematics, also demands robust reasoning to provide reliable answers, given the high standards of healthcare. However, verifying medical reasoning is challenging, unlike those in mathematics. To address this, we propose \textbf{verifiable medical problems} with a medical verifier to check the correctness of model outputs. This verifiable nature enables advancements in medical reasoning through \textbf{a two-stage approach}: (1) using the verifier to guide the search for a complex reasoning trajectory for fine-tuning LLMs, (2) applying reinforcement learning (RL) with verifier-based rewards to enhance complex reasoning further. Finally, we introduce \textbf{HuatuoGPT-o1}, a medical LLM capable of complex reasoning, which outperforms general and medical-specific baselines using only 40K verifiable problems. Experiments show complex reasoning improves medical problem-solving and benefits more from RL. We hope our approach inspires advancements in reasoning across medical and other specialized domains.
\end{abstract}

\section{Introduction}

The release of OpenAI o1 has marked a significant milestone in large language model (LLM) development, showcasing impressive capabilities \cite{guan2024deliberative,o1medicalpreliminary,o1p11}. This breakthrough highlights the potential of \textbf{scaling Chain-of-Thought (CoT)} and \textbf{reinforcement learning} to enhance LLM performance \cite{o1journey,o1p1,o1p9}.  
While subsequent research efforts attempt to replicate these advancements, they often remain limited to mathematical reasoning tasks \cite{qwq-32b-preview,luong2024reft,o1p4,o1p9}. The application of o1-like methods to specialized fields, such as medicine, remains largely underexplored.

Medical tasks often involve complex reasoning~\cite{saab2024capabilities,patel2005thinking,chen2024cod}.
In real-world medical diagnoses or decisions, doctors often deliberate carefully. Such life-critical field necessitates meticulous thinking to ensure more reliable answers \cite{o12,o1p10}. Additionally, the medical domain offers unique advantages: compared to general domains, the medical domain is generally narrower in scope and easier to verify. Furthermore, medical reasoning closely resembles real-world applications in fields like finance, law, education, and security, making advancements in this area readily transferable \cite{chen2023huatuogpt,cheng2023adapting}.

Despite these advantages, a key challenge in medical reasoning is verifying the thought process, which often lacks clear steps. Inspired by mathematical problems that allow verification through their outcomes, we construct 40K verifiable medical problems reformatted from challenging, closed-set medical exam questions. These verifiable problems are characterized as open-ended with unique, objective ground-truth answers that allow an LLM verifier to check solution correctness. This enables a two-stage approach for advancing medical complex reasoning:

\textbf{Stage 1: Learning Complex Reasoning} We construct complex reasoning trajectories through strategy-based searches guided by verifier feedback (True or False). The LLM first initializes a CoT. If the verifier rejects the current CoT, the model extends the CoT by applying a strategy sampled from \textit{Backtracking}, \textit{Exploring New Paths}, \textit{Verification}, and \textit{Correction} until a correct answer is provided. Successful reasoning trajectories  are then used to fine-tune the LLM, enabling it develop complex reasoning skills that embody iterative reflection.

\textbf{Stage 2: Enhancing Complex Reasoning with RL} After acquiring complex reasoning skills,  reinforcement learning (RL) further refine this ability. Specifically, sparse rewards provided by the verifier guide self-improvement using the Proximal Policy Optimization (PPO) algorithm.

Using this approach, we present \textbf{HuatuoGPT-o1}, a medical LLM capable of producing a long CoT to recognize its mistakes, try different strategies and refine the answer. Experiments demonstrate that our method (using only 40K data points) yields an 8.5-point improvement on medical benchmarks with an 8B model. Furthermore, our 70B model outperforms other open-source general and medical-specific LLMs across multiple medical benchmarks. The experiments further reveal that complex reasoning enhances medical problem-solving and boosts RL performance compared to standard or non-CoT methods.
Our contributions are as follows:
\squishlist
   \item To the best of our knowledge, this is the first work to advance medical complex reasoning in LLMs using verifiable medical problems and a medical verifier.

     \item With verifiable medical problems, we propose a two-stage training approach, combining search strategies to construct reasoning pathways for fine-tuning, and further enhanced by RL with verifier feedback.
 
    \item Using the proposed method, we developed HuatuoGPT-o1, the first medical LLM capable of complex reasoning. HuatuoGPT-o1 exhibits superior performance compared to the open-source general and medical-specific baselines.

    \item  Our experiments reveal that complex reasoning is effective for medical problem-solving and benefits RL enhancements.
    
\squishend

\section{Verifiable Medical Problems}

\begin{figure*}[ht!]
  \centering
  \includegraphics[width=0.95\textwidth]{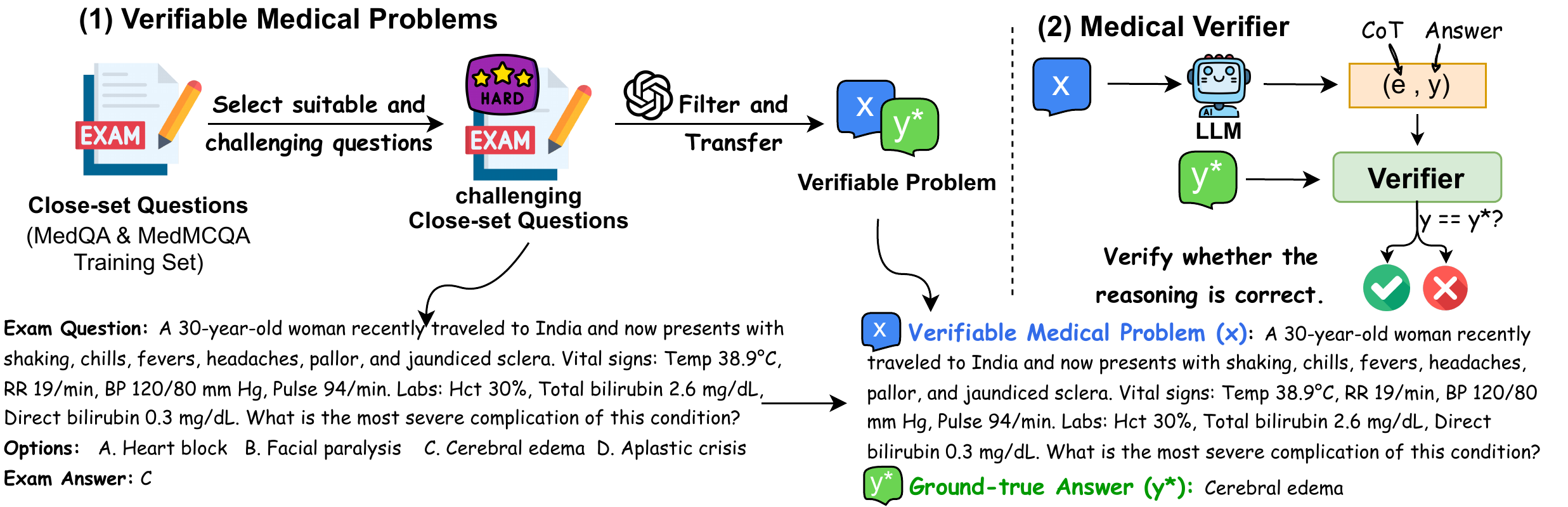}
  \caption{\label{fig2-1}\textbf{Left:} Constructing verifiable medical problems using challenging close-set exam questions. \textbf{Right:} The verifier checks the model’s answer against the \textit{ground-truth answer}.}
\end{figure*}

Inspired by mathematical problems that enable verification of the solution process through the final result, we aim to create verifiable medical problems that allow reasoning verification through outcomes. These verifiable problems are characterized as open-formal with unique, objective ground-truth answers, as illustrated in Figure \ref{fig2-1}.

\paragraph{Sourcing from Medical Exam Questions} To achieve this, we utilize closed-set real-world exam questions for two key reasons: 1) a large number of medical exam questions are available; and 2) these exam questions are typically objective and accurate. Specifically, we collected 192K medical multiple-choice exam questions from the training sets of  \textit{MedQA-USMLE} \cite{medqa} and \textit{MedMcQA} \cite{pal2022medmcqa}.

\paragraph{Transforming to Verifiable Medical Problems} However, these medical questions are closed-set, meaning they provide limited options to choose from. This makes it easy for models to guess the correct answer without proper reasoning. Additionally, some questions are not suitable due to they may lack a unique correct answer for verification or are too simple to require reasoning.

To address this, we select and process the questions as follows: 
\squishlisttwo
    \item \textbf{Selecting Challenging Questions} We removed questions that three small LLMs (Gemma2-9B \cite{team2024gemma}, LLaMA-3.1-8B \cite{llama3}, Qwen2.5-7B \cite{qwen2.5}) all answered correctly and discarded short questions to retain those requiring deeper reasoning.
    \item \textbf{Ensure Unique Answers:} We excluded questions asking for “incorrect options” or with multiple correct answers. A LLM (GPT-4o) is further employed to remove questions where the correct answer might not be unique or could be ambiguous.
    \item \textbf{Reformatting to Open-Ended Formal:} Using LLMs (GPT-4o), We reformatted each closed-set question into open-ended problem an open-ended problem \(x\) and a ground-truth answer \(y^*\), as shown in Figure \ref{fig2-1}.
\squishend

The prompt used for filtering and processing can be found in Appendix \ref{ap-converquestion}. After this filtering and processing,  we ultimately constructed a dataset of 40K verifiable medical questions denoted as \(\mathcal{D} = \{(x, y^*)\}\), where \(x\) is a verifiable problem and \(y^*\) the ground-truth answer.

\paragraph{Developing Medical Verifier}  With these verifiable problems, we propose a verifier to assess the correctness of model outputs. Given a medical verifiable problem \(x\), the model generates a Chain-of-Thought (CoT) \(e\) and a result \(y\). The verifier checks \(y\) against the ground-truth answer \(y^*\) and provides binary feedback as:

\[
\mathsf{Verifier}(y, y^*) \in \{\text{True}, \text{False}\}
\]

This feedback is essential for building a correct reasoning trajectory and improving reasoning performance. We use GPT-4o \cite{openai2023gpt4} as the verifier, prompting it to perform validation with the detailed prompt provided in Appendix \ref{ap-Verifier}.  Given the prevalence of aliases in the medical domain, exact match methods \cite{luong2024reft,gandhi2024sos} commonly applied in mathematics are impractical. Experiments in Section \ref{sec:exp-res} confirm this and demonstrate the reliability of the LLM-based verifier.

\section{Methodology}
\begin{figure*}[ht!]
  \centering
  \includegraphics[width=0.95\textwidth]{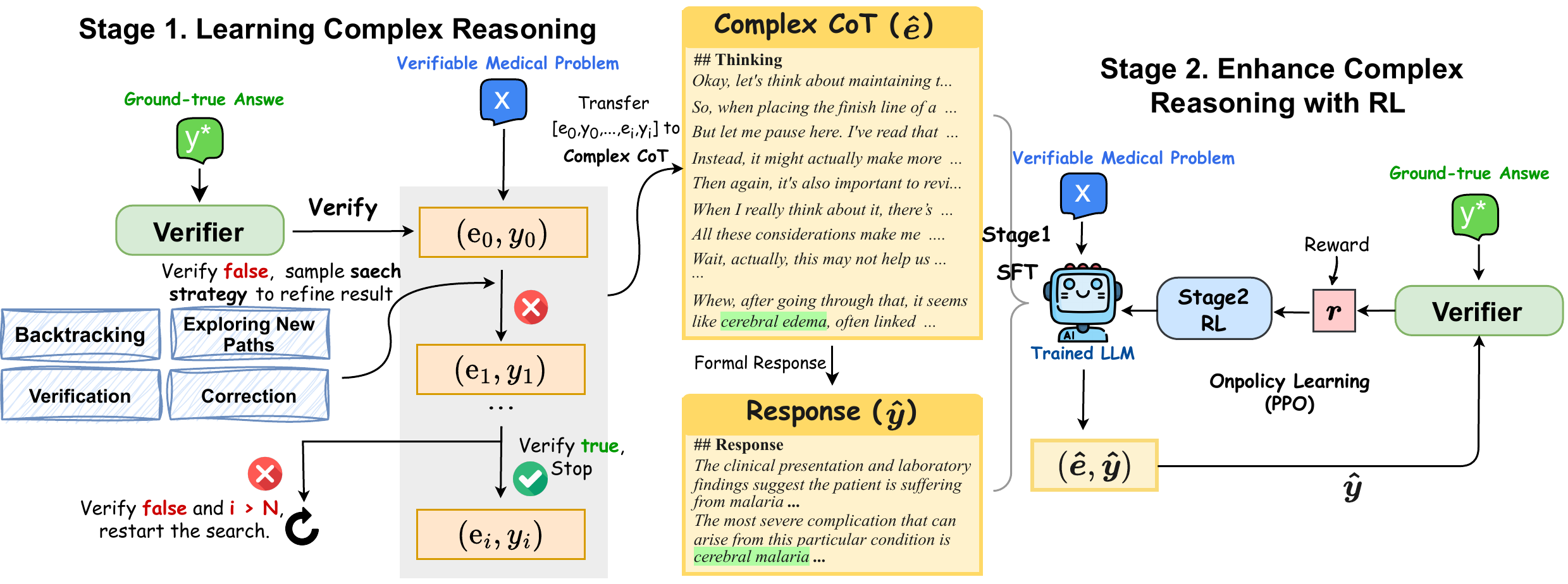}
  \caption{\label{fig3-1}Demonstration of developing and improving LLMs for medical complex reasoning. \textbf{Left (Stage1):} Searching for correct reasoning trajectories to fine-tune LLMs for complex reasoning. \textbf{Right (Stage2):} Using the verifier to enhance complex reasoning via reinforcement learning. }
\end{figure*}

In this section, we present the method for training LLMs to performing medical complex reasoning to identify errors, and refine answers using deep thinking. As shown in Figure \ref{fig2-1}, the method has two stages: \textbf{Stage One:} master complex reasoning, and \textbf{Stage Two}, enhance complex reasoning with reinforcement learning (RL).

\subsection{Stage One: Learning Complex Reasoning}

\paragraph{Searching for Correct Trajectories} Given a verifiable medical problem as a tuple \((x, y^*)\), i.e. (question, ground-true answer), the LLM (e.g., GPT-4o) generates an initial  CoT $e_0$ and answer $y_0$:
$$e_0 , y_0 = \mathsf{LLM}_{\text{init}}(x)$$

The verifier checks if $y_0$ matches $y^*$. If incorrect, the model iteratively refines the answer by applying a randomly selected search strategy $k \in \mathcal{K}$ on prior thoughts $[e_0,y_0,\ldots,e_{i-1},y_{i-1}]$, producing new reasoning  $e_i$ and new answer $y_i$:

$$ e_i , y_i =  \mathsf{LLM}_{k_i}(x,[e_0,y_0,\ldots,e_{i-1},y_{i-1}])$$
where $i$ denotes the $i$-th iteration. We define four search strategies $\mathcal{K}$  to guide the refinement process:

\squishlist

    \item \textbf{Exploring New Paths} The LLM explores a new approach $e_i$ , distinct from prior $e_0, \ldots, e_{i-1}$, to derive a new answer $y_i$.
    \item \textbf{Backtracking} The LLM revisits a previous reasoning process $e_j,y_j$, where $j< i-1$, and continues reasoning from there. Note that Backtracking is sampled only if $i \leq 2$.
    \item \textbf{Verification} The LLM evaluates the current reasoning $e_{i-1}$and result $y_{i-1}$, providing a validation process $e_i$ and the verified result $y_i$.
    \item \textbf{Corrections} The LLM critiques and corrects the current reasoning $e_{i-1}$, yielding a revised reasoning $e_j$ and answer $y_i$.    
\squishend
The process iterates until \(y_i\) is verified as correct. If the maximum iteration count  \(N = 3\) are reached, the search restarts. Each data point \((x, y^*)\) is given up to \(T = 3\) attempts; if all fail, the data point is discarded. The prompts for search reasoning trajectories can be found in Appendix \ref{ap-searchprompt}.

\begin{figure*}[t!]
  \centering
  \includegraphics[width=0.9\textwidth]{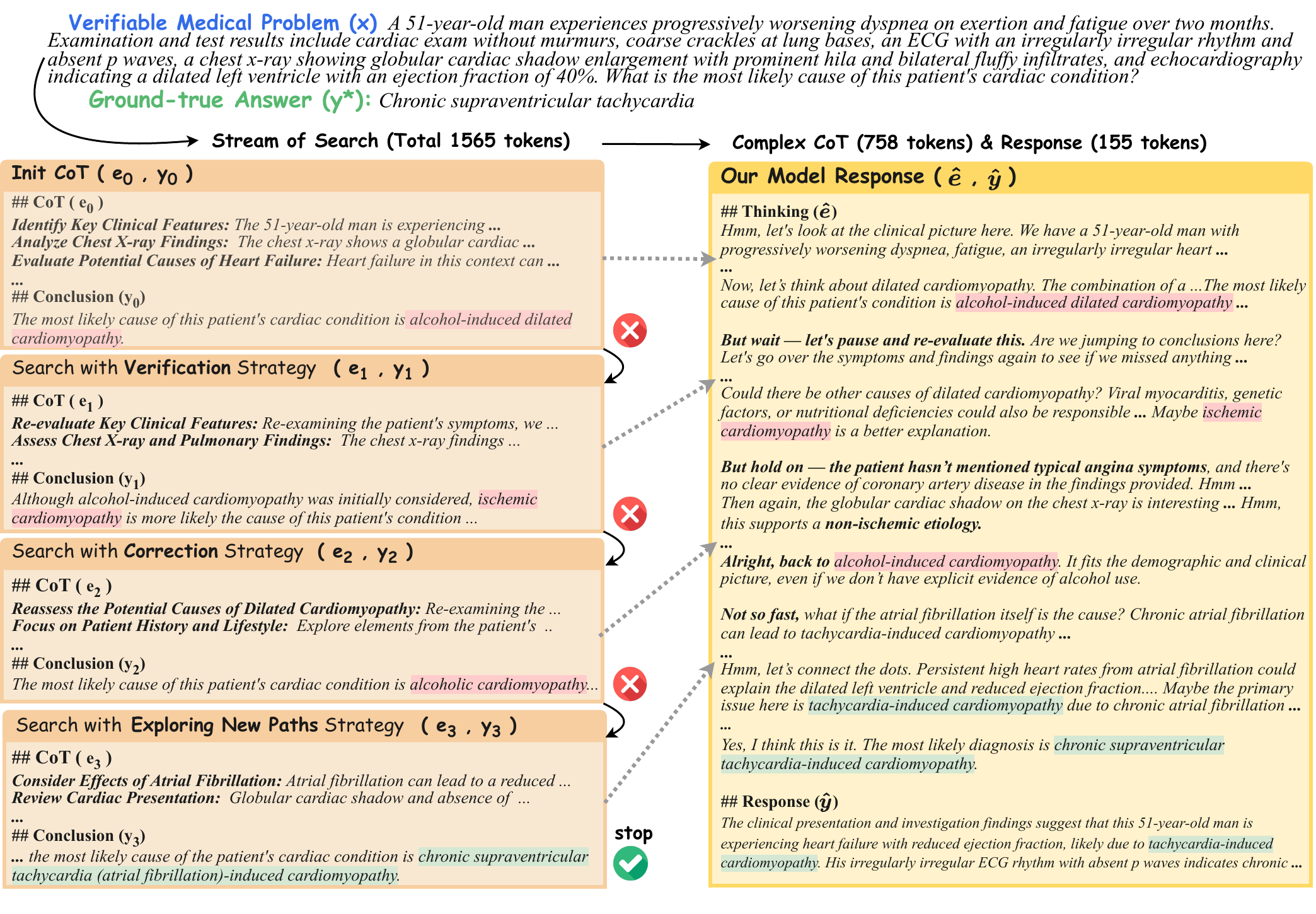}
  \caption{\label{fig3-2}Example of data synthesis. \textbf{Left:} strategy search on medical verifiable problems until the answer is validated. \textbf{Right:} Merging the entire search process into efficient complex CoTs, facilitating effective deep reasoning to refine answers. The complex CoTs and responses are used to train the model to adopt \textit{thinks-before-it-answers} behavior akin to o1.}
\end{figure*}

\paragraph{Constructing SFT Training Data} 

When a successful trajectory \([e_0, y_0, \ldots, e_i, y_i]\) is found, it is reformatted into a coherent, natural language reasoning process \(\hat{e}\) (\textit{Complex CoT}):

\[
\hat{e} = \mathsf{LLM}_{\text{Reformat}}([e_0, e_1, \ldots, e_i, y_i])
\]
As shown in Figure \ref{fig3-2}, this reformatting avoids rigid structures, using smooth transitions (e.g., “hmm,” “also,” “wait”) to streamline reasoning and reduce token usage.The model then generates a formal response $\hat{y}$ for question \(x\) using the conclusion of $\hat{e}$:

$$\hat{y} = \mathsf{LLM}_{\text{Response}}(x,\hat{e})$$

The prompt used for constructing SFT data can be found in Appendix \ref{ap-sfttrainingdata}.

\paragraph{Supervised Fine-Tuning (SFT)} We synthesize 20K SFT data points \(D_{\text{SFT}} = \{(x, \hat{e}, \hat{y})\}\) from the verifiable problem set \(\mathcal{D} = \{(x, y^*)\}\) using GPT-4o. \(D_{\text{SFT}}\) is used to fine-tune LLMs to generate a complex CoT \(\hat{e}\) followed by a formal response \(\hat{y}\). This fine-tuning process teaches the model to \textit{think before answering}, encouraging a \textit{Stream-of-Search (SoS)} \cite{gandhi2024sos} way where the model deeply explores and refines its reasoning before answering.

\begin{algorithm}[t!]
\small
\newcommand{\mycommfont}[1]{\normalfont\textcolor{black}{#1}}
\SetCommentSty{mycommfont}
\DontPrintSemicolon

\textbf{Require}: Medical Verifiable Problems $\mathcal{\boldsymbol{D}} = \{(\boldsymbol{x}, \boldsymbol{y}^*)\}$, a $\mathsf{Verifier}$, an $\mathsf{LLM}$ (GPT-4o) for synthesizing reasoning trajectories, search strategies $\boldsymbol{\mathcal{K}}$, max search depth $\boldsymbol{N}$, max search attempts $\boldsymbol{T}$, and initial policy $\boldsymbol{\pi}_{\boldsymbol{\theta}}$.

\BlankLine

$\boldsymbol{\mathcal{D}}_\text{Search}, \boldsymbol{\boldsymbol{\mathcal{D}}}_\text{RL} \gets \mathsf{Split}(\boldsymbol{\mathcal{D}})$ \\
$\boldsymbol{\mathcal{D}}_\text{SFT} \gets \emptyset$ \\

\textcolor{deepblue}{\textit{// Stage One: Learning Complex Reasoning}}\;

\For{$(\boldsymbol{x}, \boldsymbol{y}^*) \in \boldsymbol{\mathcal{D}}_\text{Search}$}{
    \For{$j \gets 1$ \KwTo $\boldsymbol{T}$}{
        $\boldsymbol{e}_0, \boldsymbol{y}_0 \gets \mathsf{LLM}_{\mathrm{init}}(\boldsymbol{x})$ \\
        
        \For{$i \gets 1$ \KwTo $\boldsymbol{N}$}{
            $\boldsymbol{k}_i \sim \boldsymbol{\mathcal{K}} $ \\
            $\boldsymbol{e}_i, \boldsymbol{y}_i \gets \mathsf{LLM}_{\boldsymbol{k}_i}(\boldsymbol{x}, [\boldsymbol{e}_0, \boldsymbol{y}_0, ..., \boldsymbol{e}_{i-1}, \boldsymbol{y}_{i-1}])$ \\
            
            \If{$\mathsf{Verifier}(\boldsymbol{y}_i, \boldsymbol{y}^*)$}{
                $\boldsymbol{\hat{e}} \gets \mathsf{LLM}_{\text{Reformat}}([\boldsymbol{e}_0, \boldsymbol{y}_0, ..., \boldsymbol{e}_{i}, \boldsymbol{y}_{i}])$ \\
                $\boldsymbol{\hat{y}} \gets \mathsf{LLM}_{\text{Response}}(\boldsymbol{\hat{e}})$ \\
                $\boldsymbol{\mathcal{D}}_\text{SFT} \gets \boldsymbol{\mathcal{D}}_\text{SFT} \cup \{(\boldsymbol{x}, \boldsymbol{\hat{e}}, \boldsymbol{\hat{y}})\}$ \\
                \textbf{break}
            }
        }
        \If{$\mathsf{Verifier}(\boldsymbol{y}_i, \boldsymbol{y}^*)$}{
            \textbf{break}
        }
    }
}

\textcolor{deepblue}{\textit{// SFT}}\;

\For{$(\boldsymbol{x}, \boldsymbol{\hat{e}}, \boldsymbol{\hat{y}}) \in \boldsymbol{\mathcal{D}}_{SFT}$}{
    $\mathcal{L}_\text{SFT}(\boldsymbol{\theta}) \gets -\log \boldsymbol{\pi}_{\boldsymbol{\theta}}(\boldsymbol{\hat{e}}, \boldsymbol{\hat{y}} \mid \boldsymbol{x})$ \\
    $\boldsymbol{\theta} \gets \mathsf{UpdateParameters}(\mathcal{L}_{SFT}(\boldsymbol{\theta}), \boldsymbol{\theta})$ \\
}

\BlankLine

\textcolor{deepblue}{\textit{// Stage Two: Enhance Reasoning with RL}}\;

$\boldsymbol{\pi}_{\mathrm{ref}} \gets \boldsymbol{\pi}_{\boldsymbol{\theta}}$ \\
\For{$(\boldsymbol{x}, \boldsymbol{y}^*) \in \boldsymbol{\mathcal{D}}_\text{RL}$}{
    $\boldsymbol{\hat{e}}, \boldsymbol{\hat{y}} \sim \boldsymbol{\pi}_{\boldsymbol{\theta}}(\boldsymbol{x})$ \\
    \textcolor{deepblue}{\textit{// Reward}}\;
    $\boldsymbol{r} \gets \mathsf{Rule}\left(\mathsf{Verifier}\left(\boldsymbol{\hat{y}}, \boldsymbol{y}^*\right)\right) - \beta \mathsf{KL}\left(\boldsymbol{\pi}_{\boldsymbol{\theta}}\left(\cdot \mid \boldsymbol{x}\right) \mid \mid \boldsymbol{\pi}_{\mathrm{ref}}\left(\cdot \mid \boldsymbol{x}\right)\right)$ \\
    $\boldsymbol{\theta} \gets \mathsf{UpdateParameters}\left(\mathcal{L}_{\mathrm{RL}}\left(\boldsymbol{x}, \boldsymbol{\hat{e}}, \boldsymbol{\hat{y}}, \boldsymbol{r}, \boldsymbol{\pi}_{\mathrm{ref}}, \boldsymbol{\pi}_{\boldsymbol{\theta}}\right), \boldsymbol{\theta}\right)$ \\
}

\Return $\boldsymbol{\pi}_{\boldsymbol{\theta}}$
\caption{\small{Training LLMs for Medical Complex Reasoning}}

\label{algo1}
\end{algorithm}

\subsection{ Stage Two: Enhance Complex Reasoning with RL}

In this stage, we further enhance the complex reasoning skills using reinforcement learning (RL). While the LLM learned successful reasoning trajectories in stage 1, these paths, derived via search, may not be optimal. On-policy learning in stage 2 aims to refine the model for better complex CoT reasoning.

\paragraph{Rewards of RL} Rewards play a crucial role in guiding the RL training target. Given a verifiable problem \(x\) and the generated response \((\hat{e}, \hat{y})\), the reward is assigned as:

\[
r'(x, \hat{y}, y^*) = 
\begin{cases} 
1 & \text{if } \mathsf{verifier}(\hat{y}, y^*) = \text{True} \\
0.1 & \text{if } \mathsf{verifier}(\hat{y}, y^*) = \text{False} \\
0 & \text{if } \hat{y} = \text{null}
\end{cases}
\]

Following \cite{riedmiller2018learning,trott2019keeping,luong2024reft}, correct answers receive a reward of 1, incorrect answers receive 0.1, and responses that lack \textit{think-before-answering} behavior receive 0. Additionally, following related works, the total reward combines this function score with the Kullback-Leibler (KL) divergence between the learned RL policy \(\pi_\theta\) and the initial policy \(\pi_{\text{ref}}\), scaled by a coefficient \(\beta\):

\[
r(x, \hat{y}, y^*) = r'(x, \hat{y}, y^*) + \beta \mathsf{KL}(\theta)
\]
to stabilize training with sparse rewards \cite{luong2024reft}.

\paragraph{Reinforcement Learning}

For RL, We use the Proximal Policy Optimization (PPO) \cite{ppo2017} algorithm with a clipped objective. The fine-tuned model serves as the policy model \(\pi_\theta\). Training is conducted on the remaining verifiable medical problems \(\mathcal{D}_{\text{RL}} = \{(x, y^*)\}\). The policy samples responses \((\hat{e}, \hat{y})\) for input \(x\), computes the reward, and updates parameters \(\theta\).

The full training process for both stages is summarized in Algorithm \ref{algo1}.

\section{Experiments}

\subsection{Experimental Setup}

\paragraph{Training Data} Finally, We constructed a 40K medical verification dataset \(\mathcal{D} = \{(x, y^*)\}\) from the training sets of \textit{MedQA-USMLE} \cite{medqa} and \textit{MedMCQA} \cite{medmcqa}. Of this, 20K is used for SFT  in stage 1 and 20K for RL in stage 2. Additionally, 4K unconverted data (close-set questions with option answers) from $\mathcal{D}$ are included to enhance generalization. In line with prior work that integrates general-domain data to support medical adaptation \cite{chen2023huatuogpt,zhang2024ultramedical}, we add 5K general verification questions sourced from \textit{MMLU-Pro} \cite{wang2024mmlupro} outside the medical-related tracks. All data were strictly screened to avoid contamination with the evaluation data using the filtering method of Med-PaLM2 \cite{singhal2023towards} (filtering overlaps of 64 consecutive characters).

\paragraph{Model Training} Using the proposed method, we train our models \textbf{HuatuoGPT-o1-8B} and \textbf{HuatuoGPT-o1-70B} based on \textit{LLaMA-3.1-8B-Instruct} and \textit{LLaMA-3.1-70B-Instruct} \cite{llama3}, respectively. In Stage 1, the models are fine-tuned on the \(\mathcal{D}_{\text{SFT}}\) for 3 epochs with a learning rate of 5e-6 and a batch size of 128. In Stage 2, we employ PPO for RL with a learning rate of 5e-7, a batch size of 128, and \(\beta\) set to 0.03. The PPO parameters are set as: 3 PPO epochs, a discount factor 1.0, a value coefficient 1.0, and a clip range 0.2.

\paragraph{Baselines} We compare our models with two type of LLMs: \textbf{1) General LLMs}: Qwen-2.5 \cite{qwen2}, LLaMA-3.1 \cite{llama3}, Gemma 2 \cite{team2024gemma}, Yi \cite{young2024yi}, Mistral \cite{jiang2023mistral}; and \textbf{2) Medical-Specific LLMs:} UltraMedical \cite{zhang2024ultramedical}, 
OpenBioLLM \cite{pal2024openbiollms}, and BioMistral \cite{labrak2024biomistral}.

\paragraph{Benchmarks} We evaluate on standard medical benchmarks: \textit{MedQA} (USMLE test set) \cite{medqa}, \textit{MedMCQA} (validation set) \cite{pal2022medmcqa}, and \textit{PubMedQA} (test set) \cite{jin2019pubmedqa}. Aditionally, we evaluated the medical sections of some challenging LLM benchmarks, including the health and biology tracks of \textit{MMLU-Pro} \cite{wang2024mmlupro}, and the genetics and molecular biology tracks of \textit{GPQA} \cite{rein2023gpqa}. Due to the limited number of \textit{GPQA} questions, we ran this evaluation 5 times and averaged the results.

\subsection{Experimental Results}

\begin{table*}[ht!] \small \centering
\adjustbox{max width=1.0\textwidth}{
\begin{tabular}{lccccccc|c} \toprule 
&\multirow{4}{*}{\begin{tabular}[c]{@{}c@{}}\textbf{MedQA}\end{tabular}} & \multirow{4}{*}{\textbf{MedMCQA}}  & \multirow{4}{*}{\textbf{PubMedQA}}  & \multicolumn{2}{c}{\textbf{MMLU-Pro}} & \multicolumn{2}{c|}{\textbf{GPQA}}  & \multirow{4}{*}{\textbf{Avg.}} \\ 
\cmidrule(r){5-6} \cmidrule(r){7-8} 
                      &  & & & \textbf{Health}  &  \textbf{Biology} & \textbf{Genetics} & 
 \begin{tabular}[c]{@{}c@{}}\textbf{Molecular} \\ \textbf{Biology}\end{tabular}          \\ \midrule
\multicolumn{9}{c}{ \bf \it $\sim$ 8B Large Language Models }  \\ 
\includegraphics[width=0.12in]{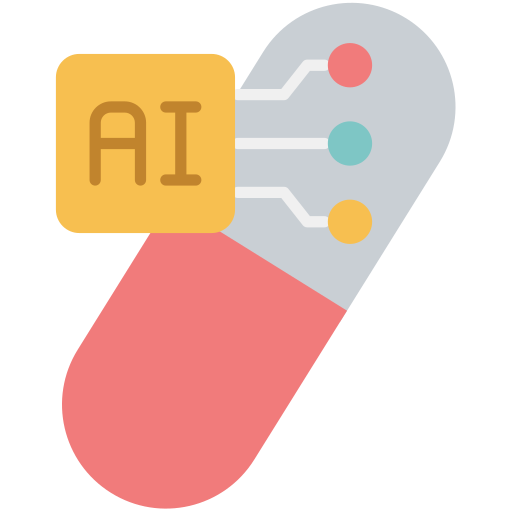} BioMistral-7B   & 45.0 & 40.2  & 66.9 & 27.4 & 49.2 & 28.6 & 38.5 & 42.3  \\
\includegraphics[width=0.12in]{pic/med2.png} OpenBioLLM-8B & 57.7 & 54.1  & 74.1 & 38.4 & 52.4 & 43.7 & 39.6 & 51.4  \\
\includegraphics[width=0.12in]{pic/med2.png}  UltraMedical-8B & \underline{71.1} & \underline{58.3}  & \underline{77.4} & \underline{55.1} & 66.7 & 41.2 & 48.4 & 59.7  \\
Mistral-7B-Instruct & 48.2 & 44.6  & 59.5 & 33.7 & 53.6 & 30.0 & 46.1 & 45.1  \\
Yi-1.5-9B-Chat & 50.8 & 48.7  & 69.8 & 43.4 & 65.6 & 42.5 & 48.1 & 52.7  \\
LLaMA-3.1-8B-Instruct & 58.7 & 56.0  & 75.2 & 52.7 & 64.6 & 33.8 & 46.8 & 55.4  \\
GLM-4-9B-Chat& 58.9 & 49.8  & 73.5 & 45.5 & 65.4 & \textbf{53.8} & 41.6 & 55.5  \\
Qwen2.5-7B-Instruct & 57.0 & 55.6  & 72.7 & 50.6 & \underline{70.2} & 36.2 & 49.7 & 56.0  \\
Gemma2-9B & 61.8 & 55.9  & 63.3 & \underline{55.1} & \textbf{74.9} & 35.0 & \underline{57.4} & 57.6  \\
\rowcolor{orange!12}\includegraphics[width=0.12in]{pic/med2.png}\includegraphics[width=0.12in]{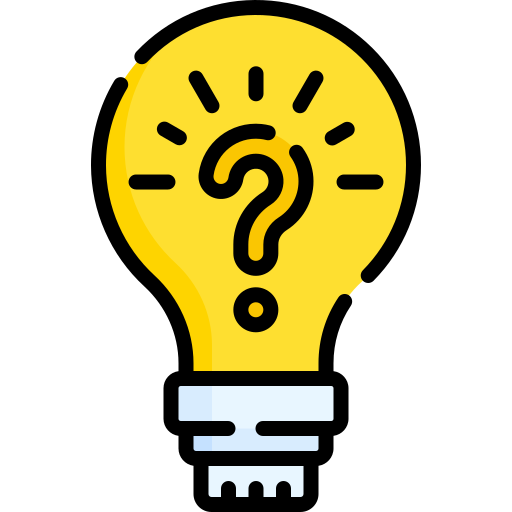} \textbf{HuatuoGPT-o1-8B} & \textbf{72.6} & \textbf{60.4}  & \textbf{79.2} &  \textbf{58.7} & 68.2 & \underline{48.8} & \textbf{59.7} & \textbf{63.9}  \\

\rowcolor{orange!12} \ \ w/o Stage2 (\textbf{\textcolor{alizarin}{RL}})  & 69.0 & 57.9  & 77.7 & 53.5 & 66.1 & 41.2 & 53.5 & \underline{59.8}  \\
\midrule
\midrule
\multicolumn{9}{c}{ \bf \it $>$ 10B Large Language Models }  \\ 
\includegraphics[width=0.12in]{pic/med2.png}  UltraMedical-70B & \underline{82.2} & 71.8  & 78.4 & 64.8 & 71.1 & 33.8 & 62.9 & 66.4  \\
\includegraphics[width=0.12in]{pic/med2.png} OpenBioLLM-70B & 76.1 & \textbf{74.7}  & \underline{79.2} & 68.8 & 76.7 & 38.8 & 54.8 & 67.0  \\
DeepSeek-67B-Chat & 57.1 & 51.7  & 76.1 & 46.9 & 66.2 & 40.0 & 51.0 & 55.6  \\
Yi-1.5-34B-Chat & 59.5 & 56.7  & 74.3 & 52.8 & 71.0 & 32.5 & 56.8 & 57.7  \\
Gemma2-27B & 65.4 & 60.2  & 72.6 & 61.1 & 76.2 & 32.5 & 61.6 & 61.4  \\
Qwen2.5-72B-Instruct & 72.7 & 66.2  & 71.7 & 65.3 & 78.8 & 41.2 & 56.8 & 64.7  \\
\includegraphics[width=0.12in]{pic/question.png} QwQ-32B-Preview & 72.3 & 65.6  & 73.7 & 62.0 & 78.1 & 37.5 & \underline{64.5} & 64.8  \\
Llama-3.1-70B-Instruct & 78.4 & 72.5  & 78.5 & 68.2 & \underline{80.8} & 52.5 & 61.6 & 70.3  \\
\rowcolor{orange!12}\includegraphics[width=0.12in]{pic/med2.png}\includegraphics[width=0.12in]{pic/question.png} \textbf{HuatuoGPT-o1-70B} & \textbf{83.3} & \underline{73.6}  & \textbf{80.6} & \textbf{71.0} & \textbf{82.8} & \textbf{56.2} & \textbf{66.5} & \textbf{73.4}  \\

\rowcolor{orange!12} \ \ w/o Stage2 (\textbf{\textcolor{alizarin}{RL}}) & 80.3 & 70.1  & 78.6 & \underline{70.2} & 79.8 & \underline{54.2} & 63.9 & \underline{71.0}  \\

\bottomrule  
\end{tabular}}
\caption{Main Results on Medical Benchmarks. LLMs with \includegraphics[width=0.12in]{pic/med2.png} are specifically trained for the medical domain, and \includegraphics[width=0.12in]{pic/question.png} indicates LLMs training for long chain-of-thought reasoning. "w/o" means "without". Within each segment, \textbf{bold} highlights the best scores, and \underline{underlines} indicate the second-best.}
\label{tab:res1}
\end{table*}

\label{sec:exp-res}
\paragraph{Main Results}  We evaluated various open-source LLMs on medical tasks, as shown in Table \ref{tab:res1}. The results indicate that prior medical-specific LLMs, like UltraMedical, excel on traditional medical benchmarks (MedQA, MedMCQA, PubMedQA) but struggle on the newer, more challenging datasets, even when the questions are medically related. This may suggest that MMLU-Pro and GPQA require not only medical knowledge but also stronger reasoning capabilities. 

Our model, HuatuoGPT-o1, performs exceptionally across all datasets. The 8B version outperforms the base model (LLaMA-3.1-8B-Instruct) by 8 points in overall evaluation. Furthermore, our 70B model surpasses other comparable open-source LLMS, including QwQ-32B, which are also developed specifically for advanced reasoning capabilities. These results demonstrate the effectiveness of our approach. Additionally, compared to only fine-tuning (\textit{w/o RL}), the two-stage training strategy significantly improves performance, benefiting from the verifiable medical problems.

\begin{table*}[ht!] \centering 
\adjustbox{max width=1.0\linewidth}{
\begin{tabular}{lcccccccc} \toprule 
& \textbf{MedQA} & \textbf{MedMCQA} & \textbf{PubMedQA} & \begin{tabular}[c]{@{}c@{}}\textbf{MMLU-Pro}\\ \textbf{(Med\includegraphics[width=0.1in]{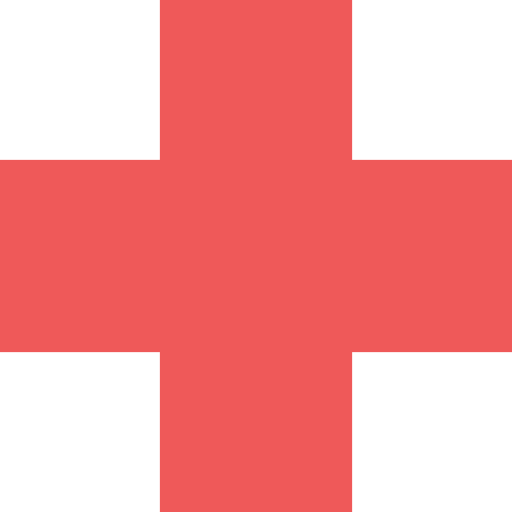})}\end{tabular}  & \begin{tabular}[c]{@{}c@{}}\textbf{GPQA}\\ \textbf{(Med\includegraphics[width=0.1in]{pic/med3.png})}\end{tabular}  \\  \midrule
\multicolumn{5}{c}{\bf \it Baseline LLMs} \\  
LLaMA-3.1-8B-Instruct & 58.7 & 56.0 & 75.2  & 58.2 & 44.1  \\ 
\midrule \midrule
\multicolumn{5}{c}{\textit{{Fine-Tuned Baseline}}} \\ 
\ \textbf{\textcolor{deepblue}{SFT}} w/ Original Exam Data of $\mathcal{D}$ & 60.0 & 55.5 & 74.1  & 54.3 & 46.9 \\ 

\midrule \midrule

\multicolumn{5}{c}{\textit{{Effectiveness of Complex Chain-of-Thought (CoT)}}} \\ 
\ \textbf{\textcolor{deepblue}{SFT}}  w/o \cancel{CoT} (only $\hat{y}$)  & 65.2 & 58.1 & 75.4  & 58.5 & 48.7  \\
 \ \textbf{\textcolor{deepblue}{SFT}}  w/ Simple CoT $(x_0, y_0)$ & 66.6 & \textbf{59.2} & 75.4  & 57.0 & 46.7   \\ 
\ \textbf{\textcolor{deepblue}{SFT}}  w/ Complex CoT $(\hat{x}, \hat{y})$ & \textbf{69.0} & 57.9 & \textbf{77.7}  & \textbf{59.4} & \textbf{51.0}  \\
\midrule \midrule
\multicolumn{5}{c}{\textit{{Effectiveness of RL}}} \\ 
\ \textbf{\textcolor{deepblue}{SFT}}  w/o \cancel{CoT} \textbf{+} \textbf{\textcolor{alizarin}{RL}} w/ PPO & 66.4 & 58.6 & 76.3  & 60.1 & 49.8  \\
\ \textbf{\textcolor{deepblue}{SFT}}  w/ Simple CoT \textbf{+} \textbf{\textcolor{alizarin}{RL}} w/ PPO  & 68.7 & 58.4 & 77.5  & 60.2 & 53.1  \\
\ \textbf{\textcolor{deepblue}{SFT}}  w/ Complex CoT \textbf{+} \textbf{\textcolor{alizarin}{RL}} w/ PPO & \textbf{72.6}  & \textbf{60.4}  & \textbf{79.2}  &  \textbf{63.1} &  \textbf{57.5}   \\ \midrule \midrule
\multicolumn{5}{c}{\textit{{Comparison of Different RL Algorithms}}} \\ 
\ \textbf{\textcolor{deepblue}{SFT}}  w/ Complex CoT \textbf{+} \textbf{\textcolor{alizarin}{RL}} w/ DPO  & 72.2  & 58.4  & 77.3  &  60.4 &  52.5  \\
\ \textbf{\textcolor{deepblue}{SFT}}  w/ Complex CoT \textbf{+} \textbf{\textcolor{alizarin}{RL}} w/ RLOO  & 71.1  & 60.1  & 78.1  &  60.9 &  58.2 \\
\ \textbf{\textcolor{deepblue}{SFT}} w/ Complex CoT \textbf{+} \textbf{\textcolor{alizarin}{RL}} w/ PPO & \textbf{72.6}  & \textbf{60.4}  & \textbf{79.2}  &  \textbf{63.1} &  \textbf{57.5}  \\
\bottomrule
\end{tabular}}
\caption{The results of ablation experiments on \textit{HuatuoHPT-o1-8B}. (Med\includegraphics[width=0.1in]{pic/med3.png}) indicates that only the medical-related parts are evaluated.  "w/o" and "w/" denote "without" and "with".  "Original Exam Data" refers to original multiple-choice questions  used for medical verifiable problems $D$. \textbf{Bold} highlights the best scores in each segment.}
\label{tab:res2}
\end{table*}

\paragraph{Ablation Study}  
We conducted an ablation study on the 8B model to analyze the impact of Complex-CoT and RL The results, shown in Table~\ref{tab:res2}, reveal the following insights:

\textbf{1. Simple Multiple-Choice Training Is Ineffective:} We compared the performance of models trained solely on the original medical multiple-choice questions of dataset \(D\). Specifically, we used multiple-choice questions as inputs and the correct option as output for fine-tuning. The results indicate that raining solely on multiple-choice questions (the fine-tuned baseline) yields minimal improvement over the base model (LLaMA-3.1-8B-Instruct). This suggests that learning correct answers alone does not improve problem-solving ability. 

\textbf{2. Effectiveness of Complex CoTs:} We further examined the impact of different types of Chain-of-Thought (CoT) reasoning. The results show that direct learning of response (\(\hat{y}\)) performs the worst, while simple CoT (\(y_0, e_0\)) offers only little benefit. In contrast, Complex CoT (\(\hat{y}_0, \hat{e}\)) significantly improves performance by an average of 4.3 points. This demonstrates the importance of teaching models to refine their answers with reflection. 

\textbf{3. Complex CoT Boosts RL:} We compared the RL enhancements under different CoT strategies, as shown in Table \ref{tab:res3}. The results indicate that Complex CoT,  which involves much longer CoT (an average of 712 tokens), yields a significantly greater gain (3.6 points) compared to simple CoT (2.6 points) and no CoT (1.1 points), as detailed in Table \ref{tab:res3}. This may suggests that longer self-play reasoning paths provide richer thought processes and feedback, enabling the model to discover higher-reward solutions.

\textbf{4. PPO Yields the Best Performance:} Using the same reward function, we further compared different RL-related algorithms, including the preference learning algorithm DPO \cite{dpo} and the REINFORCE-style algorithm RLOO \cite{rloo}. Detailed implementation information is provided in Appendix \ref{ap-rltraining}. Comparing PPO, RLOO, and DPO, we find PPO performs best, followed by RLOO and DPO. The weaker performance of DPO likely results from its off-policy nature, while PPO benefits from its use of value models, despite higher memory consumption.

\begin{table}[ht!]  \centering \small
\begin{tabular}{lcc} \toprule 
& \begin{tabular}[c]{@{}c@{}}\textbf{\# Avg. Generated }\\ \textbf{Tokens}\end{tabular}
& \begin{tabular}[c]{@{}c@{}}\textbf{$\Delta$ Avg. Gain}\\ \textbf{from RL}\end{tabular}   \\ \midrule
Direct Response (only $\hat{y}$) & 82 & 1.1 \\
Simple CoT $(x_0, y_0)$ & 281 & 2.6 \\
\rowcolor{orange!12} Complex CoT  $(\hat{x}, \hat{y})$  & \textbf{712} & \textbf{3.6} \\
\bottomrule  
\end{tabular}
\vspace{4pt}
\caption{\label{tab:res3} Comparison of models trained with different reasoning strategies. "\# Avg. Tokens" indicates the average number of tokens generated per question. $\boldsymbol{\Delta}$ represents the performance improvement from RL, as detailed in Table 1.}
\end{table}

\begin{figure}[ht!]

  \centering
  \includegraphics[width=0.50\textwidth]{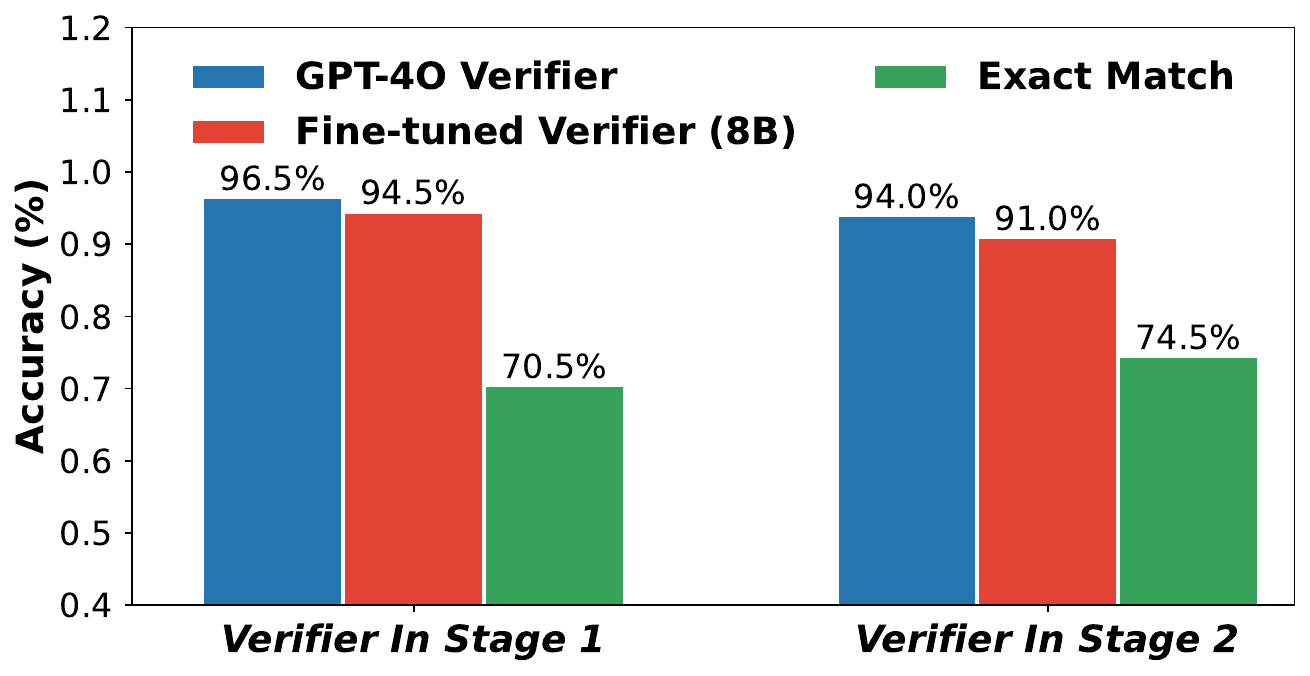}
  \caption{ \label{fig4-1} Accuracy of verifiers.  Accuracy is based on 200 manually annotated samples.}
\end{figure}

\paragraph{Reliability of the Verifier}  The verifier plays a crucial role in guiding path search and reinforcement learning (RL). In our approach, GPT-4o serves as the verifier to assess model outcomes against ground-truth answers. To assess its reliability, we manually verified 200 scoring instances sampled from Stage 1 and Stage 2. As shown in Figure~\ref{fig4-1}, GPT-4o achieved 96.5\% accuracy in Stage 1 and 94.5\% in Stage 2, demonstrating its reliability. In contrast, the Exact Match method \cite{luong2024reft}, which uses regular expressions to determine whether the correct answer is present in the response, performed significantly worse, with accuracies of only 70.5\% in Stage 1 and 74.5\% in Stage 2. This highlights the critical role of LLM-based verifiers. Additionally, we fine-tuned an 8B verifier based on LLaMA-3.1-8B with 20,000 scoring samples. The fine-tuned verifier also demonstrated feasibility, achieving over 90\% accuracy.

\begin{table}[ht!]  \centering 
\begin{tabular}{lcccccccc} \toprule  \small
&\begin{tabular}[c]{@{}c@{}}\textbf{MedQA}\\ \textbf{(Chinese)}\end{tabular} & \begin{tabular}[c]{@{}c@{}}\textbf{CMB}\\ \textbf{(Exam)}\end{tabular}  & \textbf{CMExam}  & \begin{tabular}[c]{@{}c@{}}\textbf{CMMLU}\\ \textbf{(Med\includegraphics[width=0.1in]{pic/med3.png})}\end{tabular}\\ \midrule
 \includegraphics[width=0.12in]{pic/med2.png}  HuatuoGPT2-7B & 73.7 & 63.6  & 67.4 & 58.4  
 \\
Yi-1.5-9B-Chat & 75.8 & 66.2  & 68.1 & 64.2  \\
Qwen2.5-7B & 71.4 & 70.7  & 70.4 & 70.5  \\
GLM-4-9B-chat & 75.2 & 70.0  & 70.5 & 67.6  \\
\rowcolor{orange!12}  \includegraphics[width=0.12in]{pic/med2.png}\includegraphics[width=0.12in]{pic/question.png} \textbf{HuatuoGPT-o1-7B-zh} & \textbf{79.8} & \textbf{73.0}  & \textbf{74.1} & \textbf{74.5} \\ 

\rowcolor{orange!12} \ \ w/o Stage2 (\textbf{\textcolor{alizarin}{RL}}) & 76.5 & 70.8  & 72.3 & 70.9  \\
\bottomrule  
\end{tabular}
\vspace{4pt}
\caption{\label{tab:res4}Results on Chinese medical benchmarks. \textbf{(Med\includegraphics[width=0.1in]{pic/med3.png})} indicates that only the medical portion is evaluated. MedQA (Chinese) refers to the Chinese test set of MedQA (MedQA-MCMLE).}
\end{table}

\paragraph{Domain Compatibility} To verify domain compatibility, we extra applied our method to the Chinese medical domain. We constructed a dataset of 40,000 verifiable Chinese questions from the CMB-exam training set. We then trained HuatuoGPT-o1-7B-zh using our two-stage approach based on Qwen2.5-7B-Instruct. As shown in Table \ref{tab:res4}, HuatuoGPT-o1-7B-zh outperformed other Chinese LLMs of similar size, demonstrating the method's adaptability to new domains.  For more experimental details, refer to Appendix \ref{ap-chinese}.

\section{Related Work}

\paragraph{Research on o1}  Recent studies have extensively analyzed the roadmap and core techniques of OpenAI's o1 \cite{o1journey, o1p9, o1p1}, offering foundational insights into its architecture and methodology. Extensions such as LLaMA-Berry \cite{o1p4}, LLaVA-o1 \cite{o1p8}, o1-Coder \cite{o1p5}, and Marco-o1 \cite{o1p7} have explored o1-like reasoning in various domains, including mathematics, vision-language integration, and open-ended problem-solving. However, these efforts have yet to address applications in medical or other highly specialized fields. In contrast, research focused on medicine \cite{o1medicalpreliminary, o1p2,o1p10} highlights o1's potential for deliberate, chain-of-thought reasoning in healthcare contexts. Meanwhile, several o1-inspired models, such as DeepSeek-R1-Lite-Preview \cite{bi2024deepseek}, QwQ \cite{qwq-32b-preview}, and Gemini-2.0 Flash Thinking \cite{team2023gemini}, have emerged. Despite their promise, most of these models remain closed-source, leaving substantial opportunities for further exploration and application of o1's capabilities across diverse fields.

\paragraph{Medical LLMs} The success of generalist LLMs has spurred interest in developing medical-specific LLMs to excel in the medical domain.  Notably, the MedPaLM series  \cite{singhal2023large, singhal2023towards} achieved over 60\% accuracy on the MedQA benchmark, reportedly surpassing human experts. Previous medical LLMs typically follow two main approaches \cite{zhang2024ultramedical}: \textbf{(1) Prompting Generalist LLMs} \cite{nori2023can, saab2024capabilities, li2024agent, openai2023gpt4, chen2024cod}: This method employs task-specific prompts to adapt generalist models for medical applications. 
While efficient and training-free, it is inherently limited by the capabilities of the original LLMs. \textbf{(2) Further Training with Medical Data} \cite{MedicalGPT, wang2023huatuo,han2023medalpaca, wu2024pmc, pal2024openbiollms, labrak2024biomistral, bao2023discmedllm,zhang2023biomedgpt,chen2024huatuogpt,wang2024apollo,apollomoe,christophe2024med42}: This involves training LLMs on medical pretraining corpora or medical instructions to embed medical knowledge and expertise. However, this always requires significant computational resources, such as the 1.4 billion and 3 billion training tokens used for Meditron \cite{chen2023meditron} and HuatuoGPT-II \cite{chen2023huatuogpt}. In contrast, our approach emphasizes enabling LLMs to excel in medical reasoning, offering a distinct solution.

\paragraph{Enhancing Reasoning in LLMs} Chain-of-Thought (CoT) prompting enhances the reasoning capabilities of LLMs \citep{DBLP:conf/nips/Wei0SBIXCLZ22, DBLP:conf/iclr/0002WSLCNCZ23}, but scaling expert-labeled reasoning paths remains costly, especially for complex problems \citep{DBLP:conf/emnlp/MinLHALHZ22, DBLP:journals/csur/Song0CMS23}. To mitigate this, model-generated reasoning paths filtered through external supervision offer a partial solution \citep{DBLP:conf/nips/ZelikmanWMG22, DBLP:conf/emnlp/0001GHW00023}, yet scalability challenges persist \citep{DBLP:journals/corr/abs-2305-17493, DBLP:conf/iclr/AlemohammadCLHB24}. Reinforcement learning-based methods leveraging reward models or oracle functions show potential but often suffer from slow processing, high costs, and supervision bottlenecks \citep{DBLP:conf/iclr/LightmanKBEBLLS24, DBLP:journals/corr/abs-2308-09583}.

\paragraph{Complex Reasoning}  
Developing models with reflective abilities like critique and self-correction has shown success in reasoning, planning, and coding tasks \citep{gandhi2024sos,DBLP:conf/nips/MadaanTGHGW0DPY23, DBLP:journals/corr/abs-2303-16749, DBLP:conf/iclr/WelleckLWBSK023, DBLP:conf/emnlp/XiJZZGLGZH23, DBLP:conf/eacl/PaulIPBBWF24}, though underexplored in specialized domains like medicine. 
While prompting techniques can generate self-critical reasoning \citep{bai2022constitutional, DBLP:conf/nips/MadaanTGHGW0DPY23}, they struggle without reliable reward functions or verifiers, particularly in complex domains \citep{DBLP:conf/iclr/0009CMZYSZ24, DBLP:conf/acl/XuZZP0024}. Fine-tuning and reinforcement learning methods offer solutions but require extensive human annotations or intricate reward designs \citep{DBLP:journals/corr/abs-2308-04592, DBLP:journals/corr/abs-2406-14024, DBLP:conf/iclr/ZhouWLSLQLJSZ024, DBLP:conf/icml/HavrillaRNDZHR24}. Additionally, self-training methods present a promising direction for developing self-correction capabilities \citep{DBLP:conf/iclr/WelleckLWBSK023, zheng2024criticcotboostingreasoningabilities, DBLP:journals/corr/abs-2409-12917}.

\section{Conclusion}
This study advances the medical reasoning capabilities of LLMs. Firstly, we construct the medical verifiable problems and a medical verifier. This enabled a two-stage training process: (1) learning complex reasoning and (2) enhancing it through RL.  We developed HuatuoGPT-o1, a medical LLM with \textit{thinks-before-it-answers} behavior, achieving outstanding performance in medical benchmarks. Experiments show that complex reasoning improves medical problem-solving and benefits obviously from RL.  Additional validation in Chinese medical contexts shows the method’s adaptability to other fields. We believe our approach can enhance domain-specific reasoning beyond mathematics.

\newpage

\section*{Acknowledgment}

This work was supported by  the Shenzhen Science and Technology Program (JCYJ20220818103001002), Shenzhen Doctoral Startup Funding (RCBS20221008093330065), Tianyuan Fund for Mathematics of National Natural Science Foundation of China (NSFC) (12326608), Shenzhen Key Laboratory of Cross-Modal Cognitive Computing (grant number ZDSYS20230626091302006), and Shenzhen Stability Science Program 2023. GPU devices are all supported  by the university.

{
\bibliographystyle{unsrt}
\bibliography{custom}
}


\clearpage
\appendix
\section{Ethical Statement}
Although the proposed model is a medical LLM with complex reasoning capabilities, it may still produce content that includes hallucinations or inaccuracies. Therefore, the current model is not suitable for real-world applications. Consequently, we will impose strict limitations on the use of our model. The models are not permitted for use in clinical or other industry applications where such inaccuracies could lead to unintended consequences. We emphasize the ethical responsibility of users to adhere to these restrictions in order to safeguard the safety and integrity of their applications.

\section{Constructing Medical Verifiable Problems}
\label{ap-converquestion}
To construct Medical Verifiable Problems, we begin by employing small models and rule-based methods to identify challenging questions. Subsequently, we leverage GPT-4o to perform data filtering, isolating questions that have been suitably transformed. The prompt used for this data filtering process is illustrated in Figure \ref{prompt-filter}. After selecting appropriate data, we reformat multiple-choice medical exam questions into open-ended verifiable problems using the prompt provided in Figure \ref{prompt-question}.

\begin{prompt}[title={The prompt for filtering Multiple-choice Questions}] 
{

<Multiple-choice Question>\newline
\textcolor{blue}{\texttt{\{Question\}}}\newline
\textcolor{blue}{\texttt{\{Options\}}}\newline
Correct Answer: \textcolor{blue}{\texttt{\{Answer\}}}\newline
</Multiple-choice Question>\newline

You are an expert in filtering and evaluating multiple-choice questions for advanced reasoning tasks. Your job is to evaluate a given question and determine whether it meets the following criteria:  \newline
1. **Depth of Reasoning:** The question should require deeper reasoning. If the question appears too simple, mark it as "Too Simple."\newline
2. **Unambiguous Correct Answer:** The question must have a unique and unambiguous correct answer. If the question asks for "incorrect options" or allows for multiple correct answers, mark it as "Ambiguous Answer."\newline
3. **Open-Ended Reformulation Feasibility:** The question should be suitable for reformatting into an open-ended format. If the question cannot be easily reformulated into an open-ended problem and a clear ground-truth answer, mark it as "Not Reformulatable."\newline

For each question, provide one of the following evaluations:  \newline
- "Pass" (The question meets all the criteria.)  \newline
- "Too Simple"  \newline
- "Ambiguous Answer"  \newline
- "Not Reformulatable"  
}
\end{prompt}
\captionof{figure}{\label{prompt-filter}The prompt for filtering Multiple-choice Questions. Here, \textcolor{blue}{\texttt{\{Question\}}} and \textcolor{blue}{\texttt{\{Options\}}} represents the multiple-choice question and options, and \textcolor{blue}{\texttt{\{Answer\}}} represents the correct option for the multiple-choice question.}

\begin{prompt}[title={The prompt for reformatting multiple-choice questions to open-ended verifiable problems}] 
{

I will provide you with a multiple-choice question, and your task is to rewrite it into an open-ended question, along with a standard answer. The requirements are: \newline

1. The question must be specific, targeting the point being tested in the original multiple-choice question. Ensure it is open-ended, meaning no options are provided, but there must be a definitive standard answer. \newline
2. Based on the correct answer from the original question, provide a concise standard answer. The answer should allow for precise matching to determine whether the model's response is correct. \newline

Here is the multiple-choice question for you to rewrite:\newline
<Multiple-choice Question>\newline
\textcolor{blue}{\texttt{\{Question\}}}\newline
\textcolor{blue}{\texttt{\{Options\}}}\newline
Correct Answer: \textcolor{blue}{\texttt{\{Answer\}}}\newline
</Multiple-choice Question>\newline

Please output the result in the following JSON format:\newline
```json\newline
\{\{\newline
  "Open-ended Verifiable Question": "...",\newline
  "Standard Answer": "..."\newline
\}\}\newline
```

}
\end{prompt}

\captionof{figure}{\label{prompt-question}The prompt for reformatting multiple-choice questions to open-ended verifiable problems. Here, \textcolor{blue}{\texttt{\{Question\}}} and \textcolor{blue}{\texttt{\{Options\}}} represents the multiple-choice question and options, and \textcolor{blue}{\texttt{\{Answer\}}} represents the correct option for the multiple-choice question.}

\section{The Prompt of Verifier}
\label{ap-Verifier}

GPT-4o serves as the verifier to assess the correctness of model-generated outputs. Using the prompt depicted in Figure \ref{prompt-verifier}, we present GPT-4o with both the model's output and the ground-truth answer to evaluate the correctness of the response. The verifier returns a Boolean value: \textbf{True} if the response is accurate and \textbf{False} otherwise.

\begin{figure*}[ht!]
\centering
\begin{prompt}[title={The Prompt for Verifier}] 
{
<Model Response>  \newline
\textcolor{blue}{\texttt{\{Model Response\}}}\newline
</Model Response>  \newline

<Reference Answer>  \newline
\textcolor{blue}{\texttt{\{Ground-true Answer\}}}\newline
</Reference Answer>  \newline

You are provided with a model-generated response (<Model Response>) and a reference answer (<Reference Answer>). Compare the model response with the reference answer and determine its correctness. Your task is to simply output "True" if the response is correct, and "False" otherwise.
}
\end{prompt}
\caption{The prompt for the GPT-4o verifier. \textcolor{blue}{\texttt{\{Model Response\}}} represents the output of the model to be verified. \textcolor{blue}{\texttt{\{Ground-true Answer\}}} represents the ground-truth answer for medical verifiable problems.}
\label{prompt-verifier}
\end{figure*}

\section{Prompts for Searching Trajectories}
\label{ap-searchprompt}
This section outlines the prompts used for constructing complex Chain-of-Thought (CoT) reasoning pathways. Initially, a question \(x\) is presented to GPT-4o, which generates an initial CoT response using the prompt shown in Figure \ref{prompt-initcot}. If the verifier determines the response to be incorrect, GPT-4o employs one of several search strategies to iteratively refine the output until it is accurate. The prompts for these four search strategies — \textbf{Backtracking}, \textbf{Exploring New Paths}, \textbf{Correction}, and \textbf{Verification} — are detailed in Figures \ref{prompt-newpath}, \ref{prompt-newpath}, \ref{prompt-correction}, and \ref{prompt-verification}, respectively.

\begin{prompt}[title={The prompt for initial CoT}] 
{

<question>\newline
\textcolor{blue}{\texttt{\{Question\}}}\newline
</question>\newline

Please respond to the above question <question> using the Chain of Thought (CoT) reasoning method. Your response should consist of multiple steps, each of which includes three types of actions: **"Inner Thinking"**, **"Final Conclusion"**, and **"Verification"**:\newline

- **'Inner Thinking'**: This is the step where thinking is done. Note that multiple 'Inner Thinking' steps are required to describe thorough reasoning. Each step should first generate a brief title.\newline
- **'Final Conclusion'**: At this stage, you summarize the correct reasoning from previous 'Inner Thinking' steps and provide the final answer. No title is required here.\newline
- **'Verification'**: At this stage, you verify the conclusion from the "Final Conclusion" step. If the conclusion holds, end the process. If not, return to "Inner Thinking" for further reasoning. No title is required here.\newline

The output format must strictly follow the JSON structure below: \newline
```json \newline
\{ \newline
  "CoT": [\newline
    \{"action": "Inner Thinking", "title": "...", "content": "..."\},\newline
    ...,\newline
    \{"action": "Final Conclusion", "content": "..."\},\newline
    \{"action": "Verification", "content": "..."\}\newline
  ]\newline
\}\newline
```
}
\end{prompt}
\captionof{figure}{\label{prompt-initcot}The prompt for initial CoT. \textcolor{blue}{\texttt{\{Question\}}} represents the input question, i.e., the question \(x\) of the medical verifiable problems.}

\begin{prompt}[title={The Prompt for \textbf{Backtracking} Breask Search Strategy}] 
{

<question>\newline
\textcolor{blue}{\texttt{\{Question\}}}\newline
</question>\newline

<previous reasoning>\newline
\textcolor{blue}{\texttt{\{Previous\_CoT\}}}\newline
<previous reasoning>\newline

<response requirements>\newline
Your response must include the following steps, each composed of three types of actions: **"Inner Thinking"**, **"Final Conclusion"**, and **"Verification"**:\newline

1. **Inner Thinking**: Break down the reasoning process into multiple concise steps. Each step should start with a brief title to clarify its purpose.\newline
2. **Final Conclusion**: Summarize the correct reasoning from all previous 'Inner Thinking' steps and provide the final answer. No title is needed for this section.\newline
3. **Verification**: Verify the accuracy of the "Final Conclusion". If it holds, conclude the process. Otherwise, return to "Inner Thinking" for further refinement.\newline
</response requirements>\newline

<question> represents the question to be answered, and <previous reasoning> contains your prior reasoning. Your task is to continue from the current 'Verification' step. I have manually reviewed the reasoning and determined that the **Final Conclusion** is false. Your 'Verification' results must align with mine. Proceed to refine the reasoning using **backtracking** to revisit earlier points of reasoning and construct a new Final Conclusion.\newline

\#\#\# Output Format\newline
Strictly follow the JSON structure below. You do not need to repeat your previous reasoning. Begin directly from the next 'Verification' stage.\newline

```json\newline
\{\newline
  "CoT": [\newline
    \{"action": "Verification", "content": "..."\},\newline
    \{"action": "Inner Thinking", "title": "...", "content": "..."\},\newline
    ...,\newline
    \{"action": "Final Conclusion", "content": "..."\},\newline
    \{"action": "Verification", "content": "..."\}\newline
  ]\newline
\}\newline
```

}
\end{prompt}

\captionof{figure}{\label{prompt-backtracking}The prompt for \textbf{Backtracking} search strategy. Here, \textcolor{blue}{\texttt{\{Question\}}} represents the problem \(x\) of the medical verifiable problems, and \textcolor{blue}{\texttt{\{Previous\_CoT\}}} represents the previous chain of thought process, i.e., \([e_0, y_0, \ldots, e_{i-1}, y_{i-1}]\).}

\begin{prompt}[title={The Prompt for \textbf{Exploring New Paths} Breask Search Strategy}] 
{

<question>\newline
\textcolor{blue}{\texttt{\{Question\}}}\newline
</question>\newline

<previous reasoning>\newline
\textcolor{blue}{\texttt{\{Previous\_CoT\}}}\newline
<previous reasoning>\newline

<response requirements>\newline
Your response must include the following steps, each composed of three types of actions: **"Inner Thinking"**, **"Final Conclusion"**, and **"Verification"**:\newline

1. **Inner Thinking**: Break down the reasoning process into multiple concise steps. Each step should start with a brief title to clarify its purpose.\newline
2. **Final Conclusion**: Summarize the correct reasoning from all previous 'Inner Thinking' steps and provide the final answer. No title is needed for this section.\newline
3. **Verification**: Verify the accuracy of the "Final Conclusion". If it holds, conclude the process. Otherwise, return to "Inner Thinking" for further refinement.\newline

</response requirements>\newline

<question> represents the question to be answered, and <previous reasoning> contains your prior reasoning. Your task is to continue from the current 'Verification' step. I have manually reviewed the reasoning and determined that the **Final Conclusion** is false. Your 'Verification' results must align with mine. Proceed to refine the reasoning by exploring new approaches to solving this problem and construct a new Final Conclusion.\newline

\#\#\# Output Format\newline
Strictly follow the JSON structure below. You do not need to repeat your previous reasoning. Begin directly from the next 'Verification' stage.\newline

```json\newline
\{\newline
  "CoT": [\newline
    \{"action": "Verification", "content": "..."\},\newline
    \{"action": "Inner Thinking", "title": "...", "content": "..."\},\newline
    ...,\newline
    \{"action": "Final Conclusion", "content": "..."\},\newline
    \{"action": "Verification", "content": "..."\}\newline
  ]\newline
\}\newline
```

}
\end{prompt}

\captionof{figure}{\label{prompt-newpath}The prompt for \textbf{Exploring New Paths} search strategy. Here, \textcolor{blue}{\texttt{\{Question\}}} represents the problem \(x\) of the medical verifiable problems, and \textcolor{blue}{\texttt{\{Previous\_CoT\}}} represents the previous chain of thought process, i.e., \([e_0, y_0, \ldots, e_{i-1}, y_{i-1}]\).}

\begin{prompt}[title={The Prompt for \textbf{Correction} Breask Search Strategy}] 
{

<question>\newline
\textcolor{blue}{\texttt{\{Question\}}}\newline
</question>\newline

<previous reasoning>\newline
\textcolor{blue}{\texttt{\{Previous\_CoT\}}}\newline
<previous reasoning>\newline

<response requirements>\newline
Your response must include the following steps, each composed of three types of actions: **"Inner Thinking"**, **"Final Conclusion"**, and **"Verification"**:\newline

1. **Inner Thinking**: Break down the reasoning process into multiple concise steps. Each step should start with a brief title to clarify its purpose.\newline
2. **Final Conclusion**: Summarize the correct reasoning from all previous 'Inner Thinking' steps and provide the final answer. No title is needed for this section.\newline
3. **Verification**: Verify the accuracy of the "Final Conclusion". If it holds, conclude the process. Otherwise, return to "Inner Thinking" for further refinement.\newline

</response requirements>\newline

<question> represents the question to be answered, and <previous reasoning> contains your prior reasoning. Your task is to continue from the current 'Verification' step. I have manually reviewed the reasoning and determined that the **Final Conclusion** is false. Your 'Verification' results must align with mine. Proceed to refine the reasoning by making precise **corrections** to address prior flaws and construct a new Final Conclusion.\newline

\#\#\# Output Format\newline
Strictly follow the JSON structure below. You do not need to repeat your previous reasoning. Begin directly from the next 'Verification' stage.\newline

```json\newline
\{\newline
  "CoT": [\newline
    \{"action": "Verification", "content": "..."\},\newline
    \{"action": "Inner Thinking", "title": "...", "content": "..."\},\newline
    ...,\newline
    \{"action": "Final Conclusion", "content": "..."\},\newline
    \{"action": "Verification", "content": "..."\}\newline
  ]\newline
\}\newline
```

}
\end{prompt}

\captionof{figure}{\label{prompt-correction}The prompt for \textbf{Correction} search strategy. Here, \textcolor{blue}{\texttt{\{Question\}}} represents the problem \(x\) of the medical verifiable problems, and \textcolor{blue}{\texttt{\{Previous\_CoT\}}} represents the previous chain of thought process, i.e., \([e_0, y_0, \ldots, e_{i-1}, y_{i-1}]\).}

\begin{prompt}[title={The Prompt for \textbf{Verification} Breask Search Strategy}] 
{

<question>\newline
\textcolor{blue}{\texttt{\{Question\}}}\newline
</question>\newline

<previous reasoning>\newline
\textcolor{blue}{\texttt{\{Previous\_CoT\}}}\newline
<previous reasoning>\newline

<response requirements>\newline
Your response must include the following steps, each composed of three types of actions: **"Inner Thinking"**, **"Final Conclusion"**, and **"Verification"**:\newline

1. **Inner Thinking**: Break down the reasoning process into multiple concise steps. Each step should start with a brief title to clarify its purpose.\newline
2. **Final Conclusion**: Summarize the correct reasoning from all previous 'Inner Thinking' steps and provide the final answer. No title is needed for this section.\newline
3. **Verification**: Verify the accuracy of the "Final Conclusion". If it holds, conclude the process. Otherwise, return to "Inner Thinking" for further refinement.\newline

</response requirements>\newline

<question> represents the question to be answered, and <previous reasoning> contains your prior reasoning. Your task is to continue from the current 'Verification' step. I have manually reviewed the reasoning and determined that the **Final Conclusion** is false. Your 'Verification' results must align with mine. Proceed to refine the reasoning by conducting a thorough **validation** process to ensure validity and construct a new Final Conclusion.\newline

\#\#\# Output Format\newline
Strictly follow the JSON structure below. You do not need to repeat your previous reasoning. Begin directly from the next 'Verification' stage.\newline

```json\newline
\{\newline
  "CoT": [\newline
    \{"action": "Verification", "content": "..."\},\newline
    \{"action": "Inner Thinking", "title": "...", "content": "..."\},\newline
    ...,\newline
    \{"action": "Final Conclusion", "content": "..."\},\newline
    \{"action": "Verification", "content": "..."\}\newline
  ]\newline
\}\newline
```

}
\end{prompt}

\captionof{figure}{\label{prompt-verification}The prompt for \textbf{Verification} search strategy. Here, \textcolor{blue}{\texttt{\{Question\}}} represents the problem \(x\) of the medical verifiable problems, and \textcolor{blue}{\texttt{\{Previous\_CoT\}}} represents the previous chain of thought process, i.e., \([e_0, y_0, \ldots, e_{i-1}, y_{i-1}]\).}

\section{Prompts for Constructing SFT Training Data} 
\label{ap-sfttrainingdata}
When a successful trajectory \([e_0, y_0, \ldots, e_i, y_i]\) is found, it is reformatted into a coherent, natural language reasoning process \(\hat{e}\) (\textit{Complex CoT}) using the prompt shown in Figure \ref{prompt-complexcot}. This reformatting avoids rigid structures, using smooth transitions (e.g., “hmm,” “also,” “wait”) to streamline reasoning and reduce token usage. The model then generates a formal response $\hat{y}$ for for question \(x\) using the conclusion of $\hat{e}$ with the prompt in Figure \ref{prompt-complexcot}.

\begin{prompt}[title={The prompt for reformatting a reasoning trajectory to complex CoT}] 
{

<Thought Process>\newline
\textcolor{blue}{\texttt{\{Thought\_Process\}}}\newline
</Thought Process>\newline

<Question>\newline
\textcolor{blue}{\texttt{\{Question\}}}\newline
</Question>\newline

The <Thought Process> above reflects the model's reasoning based on the <Question>. Your task is to rewrite the <Thought Process> to resemble a more human-like, intuitive natural thinking process. The new version should:\newline

1. Be presented as step-by-step reasoning, with each thought on a new line separated by a line break.\newline
2. Avoid structured titles or formatting, focusing on natural transitions. Use casual and natural language for transitions or validations, such as "hmm," "oh," "also," or "wait."\newline
3. Expand the content, making the reasoning richer, more detailed, and logically clear while still being conversational and intuitive.\newline

Return directly the revised natural thinking in JSON format as follows:\newline
```json\newline
\{\newline
  "NaturalReasoning": "..."\newline
\}\newline

}
\end{prompt}

\captionof{figure}{\label{prompt-complexcot}The prompt for reformatting a reasoning trajectory to complex CoT $\hat{e}$. Here, \textcolor{blue}{\texttt{\{Thought\_Process\}}} represents the successful reasoning trajectory of $[e_0, y_0, \ldots, e_i, y_i]$, and \textcolor{blue}{\texttt{\{Question\}}} represents the question $x$.}

\begin{prompt}[title={The prompt for generating a formal response with complex CoT}] 
{

<Internal Thinking>\newline
\textcolor{blue}{\texttt{\{Complex\_CoT\}}}\newline
</Internal Thinking>\newline

<Question>\newline
\textcolor{blue}{\texttt{\{Question\}}}\newline
</Question>\newline

The <Internal Thinking> represents your internal thoughts about the <Question>. Based on this, generate a rich and high-quality final response to the user. If there is a clear answer, provide it first. Ensure your final response closely follows the <Question>. The response style should resemble GPT-4's style as much as possible. Output only your final response, without any additional content.

}
\end{prompt}

\captionof{figure}{\label{prompt-response}The prompt for generating a formal response $\hat{y}$ with complex CoT $\hat{e}$. Here, \textcolor{blue}{\texttt{\{Complex\_CoT\}}} represents the complex CoT $\hat{e}$, and \textcolor{blue}{\texttt{\{Question\}}} represents the question $x$.}

\section{Settings of other RL training}
\label{ap-rltraining}  we further compared different RL-related algorithms with PPO. Specifically, we employed the preference-learning algorithm DPO and the REINFORCE-style algorithm RLOO.

\paragraph{DPO} For DPO, we had the model generate five answers for each question offline and used a verifier to identify pairs of one correct and one incorrect answer. If no such pairs were found, the data was discarded. Verified correct answers were used as positive examples, while failed verifications served as negative examples for training DPO. The hyperparameters for DPO training were set as follows: learning rate of 1e-6, batch size of 128, and a regularization parameter of 1.

\paragraph{RLOO} For RLOO, we used the same reward function as PPO. The parameters were also identical to those of PPO, with an additional parameter rloo\_k set to 2.

\section{Chinese Medical Model}
\label{ap-chinese}

\paragraph{Model Training} For the Chinese medical domain, we replaced the exam questions from the CMB training set for Chinese medical verifiable problems. Based on the same training process as the English version of HuatuoGPT-o1, we developed HuatuoGPT-o1-7B-zh, built on the Qwen2.5-7B-Instruct model.

\paragraph{Chinese Medical Evaluation} To assess the Chinese medical capabilities, we evaluated the model on three Chinese medical benchmarks, including the Chinese test set from MedQA (MCMLE) \cite{medqa}, the test set from CMB-Exam \cite{cmb}, and the test set from CMExam \cite{liu2024benchmarking}. Additionally, we evaluated the model on the medical section of the Chinese general evaluation benchmark CMMLU \cite{li2023cmmlu}, covering tracks of 'clinical knowledge,' 'agronomy,' 'college medicine,' 'genetics,' 'nutrition,' 'Traditional Chinese Medicine,' and 'virology'.

\paragraph{Comparison Models} We compared the performance of three general Chinese models: Qwen2.5, GLM-4, and Yi. Additionally, we included a comparison with a Chinese medical model, HuatuoGPT-2-7B \cite{chen2023huatuogpt}.


\end{document}